\documentclass[twocolumn,english]{IEEEtran}
\usepackage[T1]{fontenc}
\usepackage[latin9]{inputenc}
\usepackage{verbatim}
\usepackage{graphicx}

\makeatletter

\providecommand{\tabularnewline}{\\}

 \let\oldforeign@language\foreign@language
 \DeclareRobustCommand{\foreign@language}[1]{%
   \lowercase{\oldforeign@language{#1}}}

\makeatother

\usepackage{babel}
\begin{document}

\title{Some Theorems for Feed Forward Neural Networks}

\author{K.Eswaran, Member IEEE and Vishwajeet Singh %
\thanks{This paper has been sent for publication. K. Eswaran is with the Department of Computer Science and Engineering, Sreenidhi Institute of Science and Technology, Jawaharlal Nehru University, Yamnampet, Ghatkesar, Hyderabad, Telangana, 501301 INDIA e-mail: kumar.e@gmail.com; Vishwajeet Singh is with Altech Power and Energy Systems, Villa
Springs, Kowkur Bolarum Secunderabad,500010 INDIA e-mail: vsthakur@gmail.com%
}}
\maketitle
\begin{abstract}
In this paper we introduce a new method which employs the concept
of {}``Orientation Vectors'' to train a feed forward neural network.
It is shown that this method is suitable for problems where large
dimensions are involved and the clusters are characteristically sparse.
For such cases, the new method is not NP hard as the
problem size increases. We `derive' the present technique by starting
from Kolmogrov's method and then relax some of the stringent conditions.
It is shown that for most classification problems three layers are
sufficient and the number of processing elements in the first layer
depends on the number of clusters in the feature space. We explicitly
demonstrate that for large dimension space as the number of clusters
increase from N to N+dN the number of processing elements in the first
layer only increases by d(logN), and as the number of classes increase,
the processing elements increase only proportionately, thus demonstrating
that the method is not NP hard with increase in problem size. Many
examples have been explicitly solved and it has been demonstrated
through them that the method of Orientation Vectors requires much
less computational effort than Radial Basis Function methods and other
techniques wherein distance computations are required, in fact the
present method increases logarithmically with problem size compared
to the Radial Basis Function method and the other methods which depend
on distance computations e.g statistical methods where probabilistic
distances are calculated. A practical method of applying the concept
of Occum's razor to choose between two architectures which solve the
same classification problem has been described. The ramifications
of the above findings on the field of Deep Learning have also been
briefly investigated and we have found that it directly leads to the
existence of certain types of NN architectures which can be used as
a {}``mapping engine'', which has the property of {}``invertibility'',
thus improving the prospect of their deployment for solving problems
involving Deep Learning and hierarchical classification. The latter
possibility has a lot of future scope in the areas of machine learning
and cloud computing. \end{abstract}



\section{Introduction}

\IEEEPARstart{A}{} typical classification or pattern recognition 
problem involves a multi-dimensional feature space. Features are variables:
they can be, for example in a medical data, blood pressure, cholesterol,
sugar content etc. of a patient, so each data point in feature space
represents a patient. Data points will be normally grouped into various
clusters in feature space, each cluster will belong to a particular
class (disease), the problem is further complicated by the fact that
more than one cluster may belong to the same class (disease). The
problem in pattern recognition is how to make a computer recognize
patterns and classify them. Such tasks in the real world can be extremely
complex as there may be thousands of clusters and hundreds of classes
in a space of a hundred or more dimensions, therefore computers are
used to detect patterns in such data and recognize classes for decision
making.

The trend in the last 25 years is to use an artificial neural network
(ANN) architecture to solve such problems with the aid of computers.
One of the great difficulties that researchers have been working with
and enduring, is that there was no known method of obtaining a suitable
architecture for a given problem and therefore various configurations
of artificial neurons aligned in different arrays were tried out and
after a great deal of trials one particular architecture which suited
the present problem was finally chosen and used for pattern recognition
and classification.

However, it must be reiterated here that the theoretical basis of
a feed forward neural network, was first provided by Kolmogorov(1957)
({[}1{]}), who first showed that a continuous function of n variables:
$f(x_{1},x_{2},..x_{n})$ can be mapped to a single function of one
variable $v(u)$. His monumental discovery was improved and perfected
by others (Lorentz, Sprecher and Hect-Nielsen) ({[}2{]}-{[}6{]}) to
demonstrate that his proof proved an existence of a three layer network
for every pattern recognition problem that was classifiable. However,
though his theorem proved the existence of a neural architecture,
there was no easy way of actually constructing the mapping for a given
practical problem ({[}7{]}-{[}12{]}). To quote Hect-Nielsen, {}``The
proof of the theorem is not constructive, so it does not tell us how
to determine these quantities. It is strictly an existence theorem.
\footnote{Then he goes on to say ``Unfortunately,there does does not appear too much hope that a method for finding the Kolmogorov network will be developed soon''}  It tells us that
such a three layer mapping network must exist, but it doesn't tell
us how to find it'' ({[}13{]}-{[}15{]}).

Subsequently, Rumelhart, Hinton and Williams (1986) used the error minimization method to  implement and popularize  the Back Propagation (BP) algorithm which was discovered and developed by  many researchers from Bryson; Kelley; Ho; Dreyfus; Linnainmaa; Werbos and Speelpeening during a 22 year period 1960-82 (see the comprehensive review by J. Schmidhuber {[}18{]} for precise historical details and  references cited therein). The B.P. can be used for training artificial neural networks (ANN) consisting of multilayered processing elements (perceptrons) for solving general types of classification problems. Ever since then ANNs have been used over a very wide area of AI problems and currently in Deep Learning.Though these developments were wide and varied there was no good method of obtaining an optimal Architecture for an ANN for any particular problem.

\begin{figure}[htp]
\begin{center}
\includegraphics[scale=0.30]{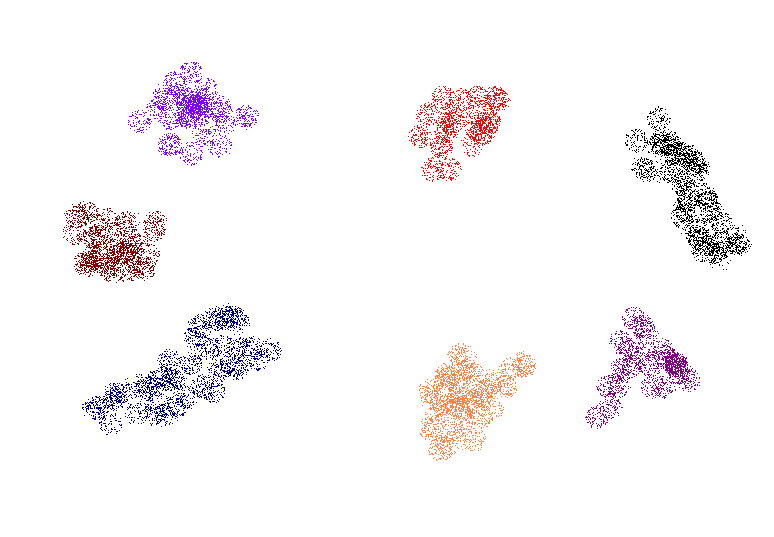} \caption{Cluster Problem for Classification}

\label{fig:fig-a} 
\end{center}
\end{figure}

For a given problem and for a given data set, various number of processing
elements and very many layers were tried out before arriving at a
particular architecture which suits the problem. However, this said,
it was however well known that a three layer architecture  is sufficient for any classification problem involving many samples which form clusters in feature space(Lippmamm, 1987, see Fig 14 in p.14 and Fig 15 p. 16 in {[}27{]}). The idea was that any group of clusters can be separated from the others by confining each
cluster inside a polygon (or polytope ) by well defined lines or planes
which form the convex hull of each cluster. See figure \ref{fig:fig-b}
below.

\begin{figure}[htp]
 \begin{center}
\includegraphics[scale=0.30]{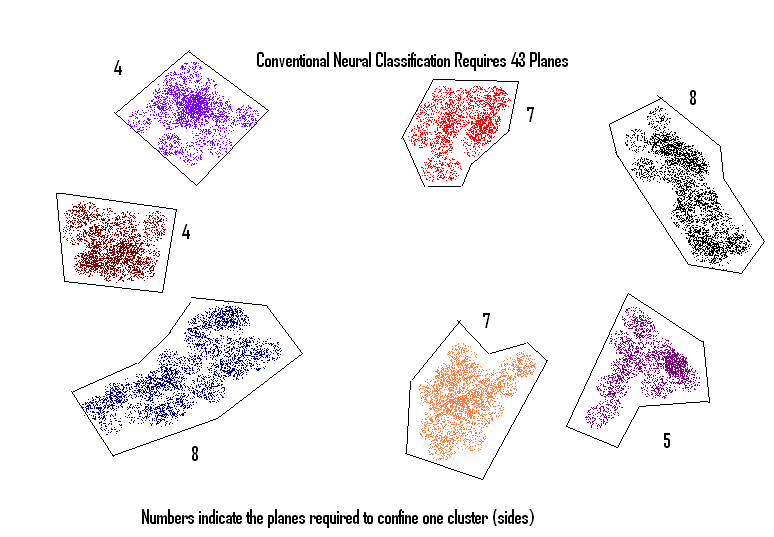} \caption{Cluster Problem for Classification solved in conventional way}

\label{fig:fig-b} 
\end{center}
\end{figure}

A classifier can then easily be constructed to discover if any point
(sample) lies within any particular polytope (cluster), all the classifier
has to do is to check if the particular point lies within the
bounding planes which circumscribe the polygon. For example, if the
polygon is a triangle, then the classifier can find out if a particular
point is within the triangle by verifying that it is within the three
sides (planes) of the triangle. If there are many polygons 
then each of them can be represented by its bounding planes (convex
hull), a classifier using ANNs can be built. By this means a three
layer ANN network wherein the first layer consists of as many processing
elements as the number of planes needed (in figure \ref{fig:fig-b} we need 43) to form the convex hulls of
all the polytopes in the sample space{[}27{]}. However, this procedure though
correct was not practically feasible, and will probably never be,
because finding the convex hull of a given region, let alone a number
of regions, is a NP hard Problem and especially so in n-dimension
feature space where the number of clusters are many and the number
of planes involved would exponentially increase. It is appropriate
to mention here that around 20 years ago there were attempts to define
the convex hull of a cluster by using sample points at the boundary
of each cluster, leading to the so called Support Vector Machines
which was introduced by Cortes and Vapnik {[}20{]}, but even these
could not get over the NP hard problem as was soon realized. (It will be seen later, how by 
adopting a new approach and the use of the concept of 'Orientation Vectors`, {[}29{]}-{[}30{]} the problems of separability and outliers are tackled, see Fig.   \ref{fig:fig-d} ). 

\begin{figure}[htp]
 \begin{center}
\includegraphics[scale=0.30]{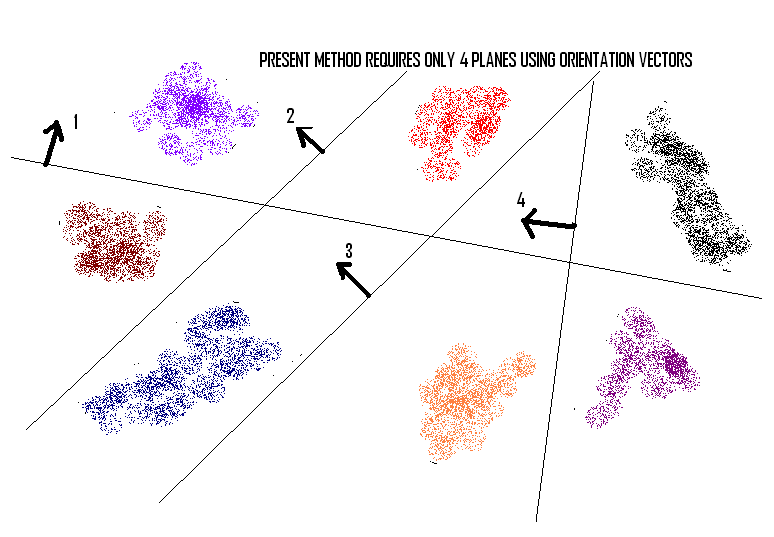}
\caption{Cluster Problem for Classification solved by present method}

\label{fig:fig-c} 
\end{center}
\end{figure}

\begin{figure}[htp]
 \begin{center}
\includegraphics[scale=0.35]{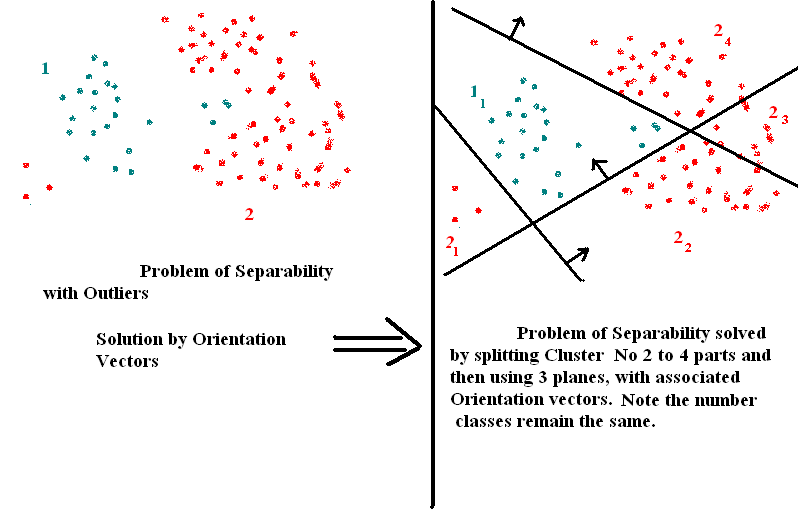} \caption{Cluster Problem for Classification with outliers}

\label{fig:fig-d}
\end{center}
\end{figure}

In this paper we describe a method to dodge the problem; we show that
it is really not necessary to find the convex hull of each cluster: 
all we require is that somehow we must be able to separate each cluster
from the other by a single plane, if this is done, then the classification
problem is much reduced and it can be performed by using a transformation
from feature space (X-space) to another S-space in such a manner that
each cluster finds itself, after transformation, in an unique quadrant
in S-space and thus each cluster is easily classified needing much less planes. See figure \ref{fig:fig-c}, where only 4 planes are necessary compared to 43 planes in figure \ref{fig:fig-b}.

The purpose of this paper is to show that this can always be done
for problems where the clusters are separable, that is if clusters
belonging to different classes do not overlap in feature space (of
course if there is an over lap there is no method which can work without
adding new features in the study - thus essentially enlarging the
dimension of the feature space). In very high dimension problems the
clusters, in practical cases, will almost always be sparse %
\footnote{If there is an image involving 30 x 30 pixels, this means we are dealing
with a 900 dimensions feature space, such a space will have $2^{900}\approx10^{270}$
quadrants; it will be hard to fill up this space even with one image
per quadrant. Thus we see that the sample points are very sparsely
distributed in feature space.%
} and this transformation from X-space to S-space performed in such
a manner that the co-domains are always within one quadrant in S-space
is not only feasible but becomes a very powerful tool for classification,
actually all the methods and results in this paper has been fashioned
from this tool.

We introduce the concept of {}``Orientation'' vectors {[}29{]}-{[}30{]} to keep track
of the clusters in feature space and solve the problem. The theorem
determines, almost precisely, the number of processing elements which
are needed for each layer to arrive at a {}``minimalistic'' architecture
which completely solves the classification problem. We further prove
that this method of classification is NOT NP hard by showing that
if the number of clusters, N, is increased then the number of processing
elements in this minimalistic architecture, at worst increases linearly
with N and at best increases by $\Delta(log(N))$.

There are two approaches that can be followed to develop the ideas
given in the previous paragraph, but eventually it so turns out that
they arrive at the same architecture, these are as follows: (i) We
follow Kolmogorov's approach but give up on the effort of exactly
trying to map the exact geometrical domain of the functions, and do
not try to obtain a continuous function that maps the exact geometrical
domains of each cluster. Instead we use piece wise continuous functions.
Further, we only make sure that we can choose planes that can separate
one domain (cluster) from another, but assume that on each of these
domains a single piece wise constant function is defined.

(ii) We give up on the idea of solving the complex hull problem for
each cluster (which is NP hard) or even on the idea of trying to confine
each cluster within planes (Lippmann 1987) which will form a polygon
(or polytope), so that all points that belong to the cluster are inside
the polytope. Instead, we just try to look for planes which separate
each cluster from another cluster by planes (half space), see figures
\ref{fig:fig-a} and \ref{fig:fig-b}. We show that if this is done
then we can solve the classification problem with much less
number of planes figure \ref{fig:fig-c}.

For the purposes of completeness and also to underline the fact that the 
ANN method is also a mapping technique and is thus related to the
Kolmogorov technique, we describe both methods (i) and (ii) in
fair detail in Section 2, though they lead to the same architecture.

In section 3, we provide a geometrical constructive proof, that under
the conditions put forth in the theorem, which demonstrates how such
a three layer network can be had if we are given the details of the
piece wise function as described above. This is one of the theorems that we prove. 

In section 4 we construct example problems wherein we use our methods. These
examples have been deliberately constructed with the intent to not only best illustrate
the method but also to show how approximations to the classification
problem can be made by assuming different ANN architectures and employ the
BP algorithm to solve it. Since these examples are so
devised  that the {}``minimalistic'' architecture %
\footnote{The term `minimalistic' that we use should be interpreted with caution,
it generally means an architecture with the minimal number of processing
elements in the 1st layer, this too is a bit imprecise: what we mean
is the number of layers when we use a sigmoid function $s_{i}=tanh(\beta y_{i})$
with $\beta$ large, say, $\beta>5$. With smaller values of $\beta$
it is possible to arrive at an architecture which has a few processing
elements less than the this `minimalistic' value, a point which becomes
clear later on.%
} for each of them is known we can compare the various approximate
with the exact solution. This study therefore gives an insight into
how one may solve practical problems when the number of clusters are
not known and the number of partitioning planes (half spaces) are
not known but have to be guessed at.

Section 5 is an application section where some suggestions as to
how one may ``guess'' the number of clusters
and the number of partitioning planes so that we can arrive
at an approximation to the 'minimalistic` architecture.

In section 6 we show that the present method is not NP hard. We end with a brief discussion on
applications on Deep Learning and a Conclusion.
7 is the conclusion.

\section{Approaches to the problem}

In this section we will detail our two approaches.

\subsection{The Kolmogorov approach}

As aforesaid, in this paper we solve a more restricted mapping problem
which is suitable for most classification tasks. We do not require
the general continuous functions of n variables $f(x_{1},x_{2},..x_{n})$,
but only require that our functions be {}``piece wise constant'',
meaning that the function$f(x_{1},x_{2},..x_{n})$ take on some constant
value in a closed set (region) (ie inside and on the boundary of a
particular cluster), therefore the function is continuous for every
point of a closed set. For example in figure \ref{fig:fig-c}: $f(x_{1},x_{2},..x_{n})$
can take on some constant value $c_{1_{1}}$ in the region $1_{1}$
and some other constant value $c_{4_{3}}$ in the region $4_{3}$,
etc. in fact there is no loss of generality if we assume that $c_{1_{1}}$
equals the class number 1 of the particular cluster $1_{1}$ ie we
may define $c_{1_{1}}=1$, similarly we may define $c_{4_{3}}=3$,
3 being the class number of cluster $4_{3}$. We further assume that
the domain of each function is separable from other domains by planes
(ie they are separable). This assumption allows us to immediately
exploit the idea of finding the minimal number of planes that can
separate the domains, thus if we know how a point (belonging to a
domain) is ``oriented'' with respect to
all the planes then we can quickly find out as to which domain the
point belongs to.

To draw a parallel with the work of Kolmogrov, we have found a way
to map a {}``piece wise constant'' function defined in n-dimension
space to a function defined in discrete 1-dimension space. A {}``piece
wise constant''' function in n-dimension space $f(x_{1},x_{2},..x_{n})$
(which takes input as points in the n-dimensional feature space and
outputs the class number of that point) is mapped to an equivalent
1-dimensional function ($v(u)$) which takes input as the cluster
number to which the point in n-dimensional feature space belongs and
outputs the corresponding class number. The function ($v(u)$) takes
as input one among a discrete set of values (cluster number) and its
set of all output values also forms a discrete set (class number).
Our construction also shows a unique way to map points in the n-dimensional
feature space to their corresponding cluster number. By this method
we use planes $S_{1}$,$S_{2}$, $S_{3}$ to separate the domains
and perform the mapping. We will see that these planes are the same
planes that are used in the second approach (next subsection), to
separate the clusters in such a manner that no two clusters are in
the same side of all the planes.

\subsection{Orientation Vector approach}

In this approach we use the concept of an `orientation vector' to
provide a geometrical constructive proof, under the conditions
put forth in the theorem, which demonstrates how such a three layer
network can be had if we are given the details of the piece wise function
as described above. 

The proof also provides a method which would overcome some
of the difficulties in arriving at a suitable architecture for a given
data in a classification problem. It is shown that given a data set
of clusters in feature space there exists an artificial neural network
architecture which can classify the data with near 100 percent accuracy
(provided the data is consistent and the train samples describe a
convex hull of each cluster). Further, it is shown how by using the
concept of an {}``orientation vector'' for each cluster, an optimal
architecture is arrived at. It is also shown that the weights of the
second hidden layers are related to the orientation vector thus making
the classification easily possible.

We give 3 examples on the method each of increasing complexity. The
purpose of these examples is to show how once the architecture of
the network is fixed, the weights of the network can be easily obtained
by using the Back Propagation algorithm to a feed forward network.

\section{Statement of Theorem and Proof}

\begin{figure}[htp]
 \begin{center}
\includegraphics[scale=0.30]{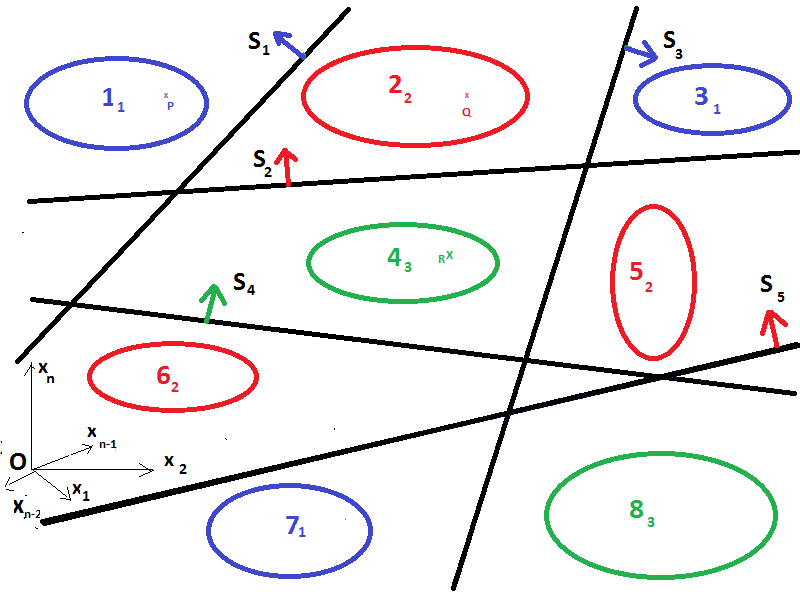} \caption{Cluster of sample points in n-dimensional space}

\label{fig:fig-e} 
\end{center}
\end{figure}

Suppose there are $m$ clusters of points in n dimensional feature
space, figure \ref{fig:fig-e} is a typical depiction, such that each
cluster of points belongs to one of $k$ distinct classes, and further
if there exist $q$ distinct $n-$dimensional planes which separate
each cluster from its neighbors in such a manner that no two clusters
are {}``on the same side'' of all the planes, then it is possible
to classify all the clusters by means of a feed forward neural network
consisting of three hidden layers which has an architecture indicated
as: $q-m-k$. That is the the neural network will have $q$ processing
elements in the first hidden layer, $m$ processing elements in the
second layer, and $k$ processing elements in the last layer. The
input to this neural net work will be $n-$dimensional, that is it
will be the coordinates of a data point in $n-$dimensional space
whose membership to a particular class will be ascertained uniquely
by this neural network. The out put of this network will be $k$ binary
numbers, out of which only one of them will be $1$ and the rest will
be zero. In the k-dimensional output vector if the the first component
is $1$ then it means the input vector belongs to the first class,
or if the second component is $1$ it means that the input vector
belongs to the second class ....and so on to the last (kth class ).
In some notations which include the number of inputs, then the architecture
would be denoted as: $n-q-m-k$.

In the figure \ref{fig:fig-e}, we have chosen: q=5, m=8 and k=3,
there are 8 clusters number 1 to 8, the suffix indicates the class
assignment, for example cluster number 4 is denoted with a suffix
3 ie as $4_{3}$ this just indicates that the points in this cluster
belong to class 3, it may be noticed that there are other points in
cluster number 8 which also belong to class 3.

\textbf{NOTE A:} It is assumed that the $q$ planes do not intersect
any cluster dividing it into two parts, if there happens to be a particular
cluster which is so divided i.e. if there is a plane which cuts a
particular cluster into two contiguous parts, then the part which
is on one side of this plane will be counted as a different cluster
from the one which is on the other side : that is the number of clusters
will be nominally increased by one : $m$ to $m+1$.

\textbf{NOTE B:} It may be noticed that we assume that all points
in a cluster belong to a single class (though the same class may be
spread to many clusters, this assumption is necessary else it means
that the n-features are not enough to separate the classes and one
would require more features. Example suppose there is a sample point
R which actually belongs to class 2 inside the cluster $4_{3}$ ,
this means class 2 and class 3 are indistinguishable in our n-dimensional
feature space and there should be more features added, thus increasing
n. The proof of the theorem of course assumes that the dimension,
n, of the selected feature space is sufficient to distinguish all
the k classes.

\textbf{NOTE C:} Perhaps it is superfluous to caution that figure
\ref{fig:fig-e} is a pictorial representation of n-dimensional space
and a plane which is merely indicated as a single line is actually
of n-1 dimensions and the arrow representing the normal (out of the
plane) is perpendicular to all these n-1 dimensions.

\begin{figure}[htp]
 \begin{center}
\includegraphics[scale=0.25]{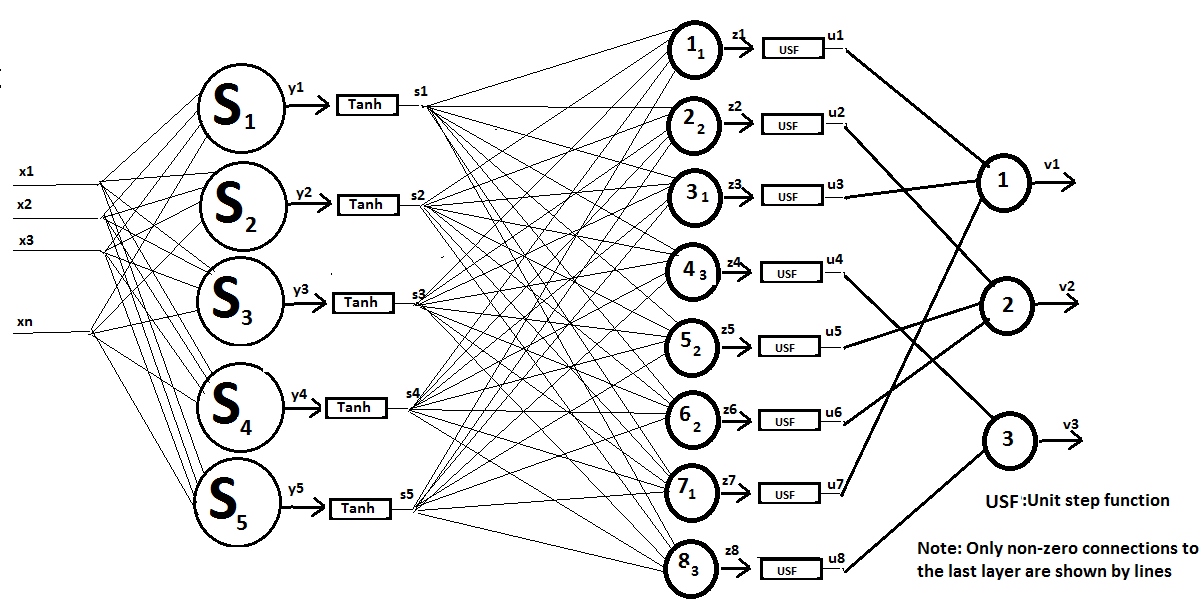} \caption{Neural network architecture proposed in this paper}

\label{fig:fig-f}
\end{center}
\end{figure}

We prove our theorem by explicit construction. To fix our notation
we have provided a diagram which shows the architecture of a typical
neural network shown in the figure \ref{fig:fig-f}, whose architecture
is chosen for classifying the clusters given in figure \ref{fig:fig-e}.

However before proceeding to the proof we need a few definitions:

We first define what is meant by the terms {}``positive side'' and
{}``negative side'' of a plane. We indicate by an arrow the normal
direction of each plane $S_{1},S_{2},...S_{q}$ (in the figure we
have taken $q=5$). It may be noticed from figure \ref{fig:fig-e}
that all the points in the particular cluster indicated by $1_{1}$
is on the side of the arrow direction of plane $S_{1}$, hence we
say that $1_{1}$ lies on the {}``positive side'' of plane $S_{1}$,or
as {}``+ve side'' of plane $S_{1}$, points on the other side of
this plane is defined to lie on the {}``negative side'' of plane
$S_{1}$ or as {}``-ve side''. So we see each cluster will be either
on the positive side or on the negative side of each plane, because
by assumption no plane cuts through a cluster, (See Note A). Now we
define a {}``orientation vector'' of a cluster as follows: Let us
take the cluster indicated by $1_{1}$ we see that this cluster is
on the +ve side of $S_{1}$, +ve side of $S_{2}$, -ve side of $S_{3}$,
+ve side of $S_{4}$ and +ve side of $S_{5}$, this situation is indicated
by the array $(1,1,-1,1,1)$. We thus introduce the concept of a orientation
vector of the cluster $1_{1}$ as a vector which has $q$ components
and is defined as $\underline{d}^{(1_{1})}=(1,1,-1,1,1)$. To take
another example, let us take the cluster $4_{3}$ its orientation
vector is $\underline{d}^{(4_{3})}=(-1,-1,-1,1,1)$ as can be ascertained
from the figure. So we can, in general denote the orientation vector
of any cluster $b$ as $\underline{d}^{b}=(d_{1}^{b},d_{2}^{b},...d_{r}^{b},..d_{q}^{b})$;
where $d_{r}^{b}$ = +1 or -1 according as cluster $b$ is on the
+ve or -ve side resp. of plane $r$. It should be noted that the orientation
vector of each cluster is unique and will not be exactly equal to
the orientation vector of another cluster; this will always happen
if the orientation vectors for each cluster are properly defined.
Thus the dot product $\underline{d}^{b}.\underline{d}^{c}$ of the
orientation vectors vectors of two different clusters $b$ and $c$
will always be less than $q$:

$\underline{d}^{b}.\underline{d}^{c}=q$ , if $b=c$

and

$\underline{d}^{b}.\underline{d}^{c}<q$ , if $b\ne c$

actually it is because of the above property and the uniqueness of
each orientation vector, that we are able to build the architecture
for any given problem.

The out put of the first processing element in the first layer denoted
by $S_{1}$ in the figure is: $s_{1}=tanh(y_{1})$

\[
s_{1}=tanh(\beta y_{1})
\]

where we arbitrarily choose $\beta=5$

\[
y_{1}=w_{10}+w_{11}x_{1}+w_{12}x_{2}+....+w_{1n}x_{n}
\]

it may be noted that the formula $w_{10}+w_{11}x_{1}+w_{12}x_{2}+....+w_{1n}x_{n}=0$
corresponds to the equation of the plane $S_{1}$ of figure \ref{fig:fig-f}.

we have similar formulae for all the processing elements, $S_{j},j=1,2,...q$
in the first layer, the last $q^{th}$ being:

\[
s_{q}=tanh(\beta y_{q})
\]

where

\[
y_{q}=w_{q0}+w_{q1}x_{1}+w_{q2}x_{2}+....+w_{qn}x_{n}
\]

where $w_{q0}+w_{q1}x_{1}+w_{q2}x_{2}+....+w_{qn}x_{n}=0$ is the
equation to the plane $S_{q}$ of figure \ref{fig:fig-f}. Since the
planes $S_{1},S_{2},..S_{q}$ are assumed to be given, the coefficients
(weights) $w_{ij}$ of all the processing elements in the first layer
are all known.

\textbf{NOTE D:} Now we wish to make a very important observation,
which has a bearing on the many things that we will be dealing with.
It may be noticed that the immediate out put of the first layer viz
$(y_{1},y_{2},..,y_{q})$ passes through the sigmoid functions $tanh(\beta y_{i})$
(with $\beta$ large, say, $\beta=5$)and then produces a vector $(s_{1},s_{2},..,s_{q})$.
But since the sigmoid function $s_{i}=tanh(\beta y_{i})$ maps almost
all the points $y_{i}$ (which are bit far away from $y_{i}=0$) to a point
close to either $s_{i}=-1$ or $s_{i}=+1$, we see that as a consequence
that every input sample $(x_{1},y_{2},..,x_{n})$ maps to $(s_{1},s_{2},..,s_{q})$
where $(s_{1}\approx\pm1,s_{2}\approx\pm1,..,s_{i}\approx\pm1,..,s_{q}\approx\pm1)$
that is the image point in S-space is always close to some q-dimensional
Hamming vector which is $(\pm1,\pm1,..,\pm1)$ .

\textbf{Definition of the {}``Center'' of a Quadrant in S-space:}
Suppose a point Q has a coordinates which can be expressed as a Hamming
Vector say $(1,-1,1,..,1)$, then we consider Q as the {}``Center''
of that quadrant of the space whose points are having coordinates:
$(s_{1}>0,s_{2}<0,s_{3}>0,..,s_{q}>0)$. Eg (i) The point Q' whose
coordinates are $(1,1,..,1)$ is the ``Center''
of the {}``first'' quadrant whose points are having coordinates:
$(s_{1}>0,s_{2}>0,s_{3}>0,..,s_{q}>0)$, similarly Eg. (ii)the point
Q'' whose coordinates are: $(-1,-1,-1,..,-1)$ is the ``Center''
of the last'' quadrant whose points are
having coordinates: $(s_{1}<0,s_{2}<0,s_{3}<0,..,s_{q}<0)$ . since,
we are in q-dimensional space there are $2^{q}$ quadrants, this provides
an upper limit to the number of clusters that can be separated by
q planes viz. $2^{q}$ .

So a worthwhile observation to make is that the images of all the
points which are not near any dividing plane $S_{i}$ will be points
close the Center of some quadrant in S -space. Further, all points
belonging to a particular cluster get mapped to a region very close
to the center of a particular quadrant, in other words all the images
of one particular cluster will be found near the center of its own
quadrant in S-space. (We will see later that this last property makes
it easy to employ a further mapping if we wish).

Since all points belonging to a single cluster gets mapped to its
unique quadrant in S-space, (uniqueness certainly follows because
of the uniqueness of each orientation vector, that is when the planes
are well chosen) and the we can easily classify them by {}``collecting''
the points in each quadrant. This is done by writing down the appropriate
weights of the processing elements in the second layer, which we call
the {}``Collection Layer'', because of its function. Let us start
with the first one which is shown as $1_{1}$ in the figure \ref{fig:fig-f}.

The output

\[
u_{1}=tanh(\beta z_{1})
\]

where

\[
z_{1}=w'_{10}+w'_{11}s_{1}+w'_{12}s_{2}+....+w'_{1q}s_{q}
\]

Now since we wish the the first processing element to output $u_{1}=+1$
if the input n-dimensional vector $x_{1},x_{2},...,x_{n}$ belongs
to the cluster $1_{1}$ and to out put $u_{1}=-1$ if it belongs to
any other cluster, we see that this condition will be adequately satisfied
if we choose:

\[
(w'_{11},w'_{12},w'_{13},...,w'_{1q})=(d_{1}^{1_{1}},d_{2}^{1_{1}},d_{3}^{1_{1}},..,d_{q}^{1_{1}})
\]

which we write in short hand as: $\underline{w'}_{1}=\underline{d}^{1_{1}}$

and the constant term: $w'_{10}=\frac{1}{2}-q$ . It can be easily
seen that if the sample point $x_{1},x_{2},...,x_{n}$ belongs to
the cluster $1_{1}$ then $z_{1}=1/2$ and hence the out put $u_{1}=tanh(\beta z_{1})$
becomes very close to +1, else the out put becomes very close to -1.

Similarly the second processing element indicated as $2_{2}$ will
output

\[
u_{2}=tanh(\beta z_{2})
\]

where

\[
z_{2}=w'_{20}+w'_{21}s_{1}+w'_{22}s_{2}+....+w'_{2q}s_{q}
\]

and if we choose

\[
(w'_{21},w'_{22},w'_{23},...,w'_{2q})=(d_{1}^{2_{2}},d_{2}^{2_{2}},d_{3}^{2_{2}},..,d_{q}^{2_{2}})
\]

ie. $\underline{w'}_{2}=\underline{d}^{2_{2}}$

and the constant term: $w'_{20}=\frac{1}{2}-q$

We can now write down the general term viz the output of the ith processing
element belonging to the jth class:

\[
u_{i}=tanh(\beta z_{i})
\]

where

\[
z_{i}=w'_{i0}+w'_{i1}s_{1}+w'_{i2}s_{2}+....+w'_{iq}s_{q}
\]

and if we choose

\[
(w'_{i1},w'_{i2},w'_{i3},...,w'_{iq})=(d_{1}^{i_{j}},d_{2}^{i_{j}},d_{3}^{i_{j}},..,d_{q}^{i_{j}})
\]

ie. $\underline{w'}_{i}=\underline{d}^{i_{j}}$

and the constant term: $w'_{i0}=\frac{1}{2}-q$

It can be easily seen that if the sample point $x_{1},x_{2},...,x_{n}$
belongs to the cluster $i_{j}$ then $z_{i}=1/2$.

\subsection{Use of Unit Step Function}

Now to simplify the proof we will use the Unit Step Function instead
of the activation tanh function in the second layer, ie. instead of
defining $u_{i}=tanh(\beta z_{i})$ as above, we use the unit step
function $Usf(z)$, which we define as:$Usf(z)=1$ if $z>0$ else
$Usf(z)=0$. The output $u_{i}=Usf(\beta z_{i})$ becomes 1 or 0 (binary).

This we do only to demonstrate the proof, but in actuality in practical
cases the original tanh function would suffice with some appropriate
changes in the equations below.

Now we come to the last layer:

This is easy to do, we choose the connection weight between the processing
element in the second layer say $i_{j}$ and processing element $l$
of the last layer as equal to the Kronecker delta $\delta_{jl}$ ;(by
definition $\delta_{jl}=1,$if $j=l$ else $\delta_{jl}=0$ ).

Thus we explicitly write

\[
v_{l}=p_{l0}+p_{l1}u_{1}+p_{l2}u_{2}+..+p_{li}u_{i}+..+p_{lm}u_{m}
\]

where we now define

\[
p_{l0}=0
\]

and we choose $p_{li}=1$ if the cluster $i_{j}$ belongs to class
$l$, that is $j=l$, else we define $p_{li}=0$

With the above choices all the weights in the network are now known,
thus completely defining the Neural Network which can classify all\emph{
the data}.

To demonstrate why it works let us consider, in figure \ref{fig:fig-f},the
connection weight between the processing element $1_{1}$ and the
processing element 1 (in the last layer)$p_{11}=1$, but the connection
weight to 2 $p_{21}=0$. Similarly, the connection weight between
the processing element indicated as $6_{2}$ and 1 $p_{16}=0$, the
connection weight between $6_{2}$ and 2, $p_{26}=1$ and the connection
weight between $6_{2}$ and 3, $p_{36}=0$; thus ensuring that if
the input point $(x_{1},x_{2},..,x_{n})$ belongs to the cluster $6_{2}$
then it will be classified as class 2. We thus see that the neural
network will out put a point belonging to a cluster $i_{j}$ to the
class j as required, since the jth processing element outputs $v_{j}$
a number which is equal to 1 as the final output; the other $v_{k},k\ne j$
will out put 0. QED.

\subsection{Four layer problem}

\begin{figure}[htp]
 \begin{center}
\includegraphics[scale=0.35]{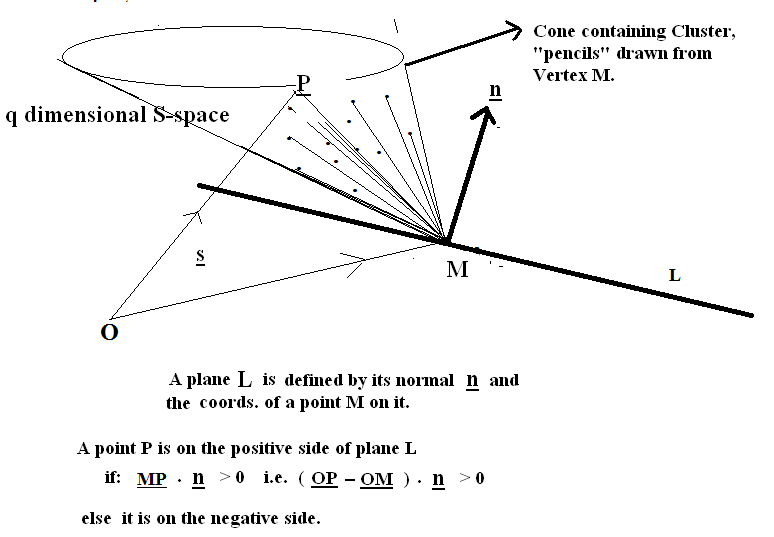} \caption{Cluster within a cone}

\label{fig:fig-g} 
\end{center}
\end{figure}

We have seen that the out put of the first layer maps all points onto
S-space; and since each cluster is mapped to its {}``own quadrant''
in this space the problem has already become separable. It was only
necessary to identify the particular quadrant that a sample had got
mapped to, in order that it can be classified; a task undertaken by
the collection layer. Though the above section shows that the number
of layers (three) is sufficient, it is sometime better to make one
more transformation from the S space to h space by using orientation
vectors in this space (see figure \ref{fig:fig-g}) this could lead
to a network with with less processing elements in the layer.

(That is if there are clusters belonging to the same class in one
half space then it is not necessary to separate these clusters individually
since anyway they belong to the same class, we can save on the number
of planes if we group such clusters as belonging to a single region.)
The figures, show how such regions, containing {}``clusters of clusters''
belonging to the same class can be separated by planes. In this section
we show how all this can be done by introducing another layer before
the ``Collection Layer''. Also the Collection
Layer in this case collects samples belonging to one region which
has samples belonging to possibly more than one cluster but all belonging
to the same class. This becomes apparent in the figures which depict
the orientation vectors H, in the s-space.

\begin{figure}[htp]
 \begin{center}
\includegraphics[scale=0.35]{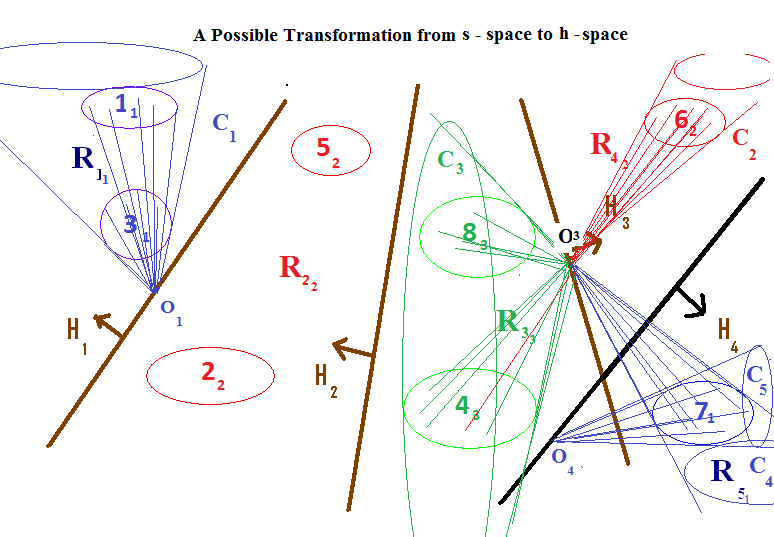} \caption{Clusters in conical pencils}

\label{fig:fig-h} 
\end{center}
\end{figure}

The above diagram shows that there are several places where clusters
belonging to the same class can be grouped as one region containing
a{}``cluster of clusters'', these regions can then be separated
by fewer planes (the figure \ref{fig:fig-h} shows 4 planes and to
prevent clutter the cone containing Region $2_{2_{2}}$ in the negative
side of $H_{1}$ and separated by the positive side of $H_{2}$ has
not been drawn).

\begin{figure}[htp]
 \begin{center}
\includegraphics[scale=0.25]{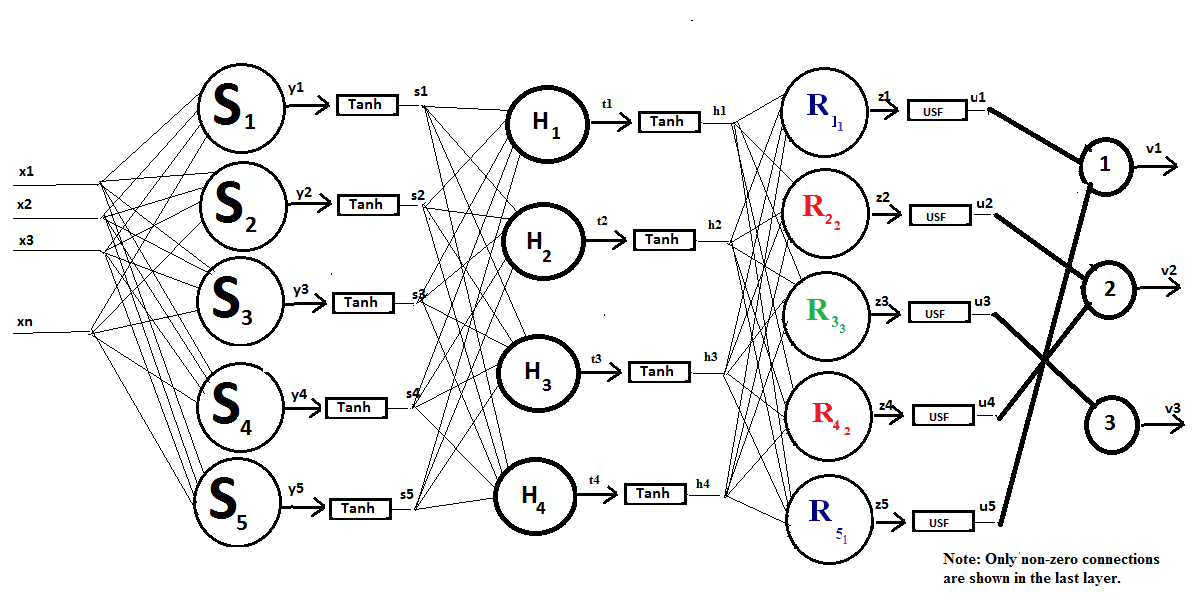} \caption{Neural network architecture of a four layer network}

\label{fig:fig-i}
\end{center}
\end{figure}

The Architecture for such a situation can be easily arrived at by
introducing the orientation vectors H as another layer . We than have
the architecture shown in figure \ref{fig:fig-i}.

$ $

In the example problems below, we did not use the Unit Step function
but rather used the Tanh function to enable the use of Back propagation
algorithm ({[}16{]}). So the outputs will be in the range $[-1,+1]$,
and all positive points are treated as +1 and all the negative points
are treated as 0.

\section{EXAMPLES}

\label{examples}

We now, for the purpose of illustration, solve a few problems and
show that for a given classification problem if a neural network architecture
is chosen as per the above theorem then the classification is guaranteed
to be 100\% correct, provided the data satisfies all the conditions
of the theorem.

\textbf{A  NOTE On Choice of Examples:} These examples , were purposely constructed because we need to know exactly the geometrical configuration of each cluster and the number of points involved and how the clusters are nested one within an another. So even though the example may seem artificial and the first one seems to be a "toy" example, they were purposely constructed so that we can theoretically calculate the minimum number of planes require to separate the clusters. This latter information is very important to us otherwise there is no way of comparing our results with the "exact" result. However, the examples become increasingly complex in Sec IV E we have r-levels of nested clusters one within another in n-dimensions, and yet they are so constructed that we know the number of planes that will separate them!

First the 3 layer architecture is chosen as per the configuration
dictated by the theorem, then the Back Propagation algorithm is used
to show that in each of the 3 example problems the classification
is 100\%.

Secondly, we introduce a fourth layer; since we cannot know the number
of processing elements in the second and third layers, different configurations
were tried. This is just to show that even if we do not know the exact
number of clusters we can by a judicious guesses choose processing
elements in the second and third layer in such a manner that the classification
is done 100\% or near 100\%.

The 3 Examples (below) are constructed in such a way that we know
the number of clusters, the number of classes and also which cluster
belongs to which class. For convenience we assume the shape of all the clusters, in the examples, to be spherical. In all the examples we generate the coordinates of sample points within clusters and coordinates of test points by using random number generators. We then use the back-propagation method after
choosing the appropriate architecture as dictated by the above theorem
for the 3 layer case and a variety of architectures for the 4 layer
case. (Later, in sections V(B) and V(D), we give techniques of choosing a suitable architecture).  
The 3 examples are: (A) The 3D cube, (B) The 4D cube and (C) The 4D nested cubes.  We then generalize in para (D) the nested cubes to n dimensions and in para (E) consider r levels of nestings in n-dimensions and give a minimalistic architecture for these and draw interesting comparisons with Radial Basis Function classifiers.

\textbf{TEST RESULTS ON EXAMPLES:}

\subsection{Three Dimensional Cube}

This is a 3-d problem involving a cube which is centered at origin
and whose neighboring vertices at a distance of 2 apart from each
other. There are 8 clusters, centered at each of the vertices, we
assume that each cluster has a radius of 0.3. The symmetrically opposite
vertices of the cube belong to the same class, and hence there are
a total of 4 classes. For example, the symmetrically opposite vertex
of (1, -1, 1) is (-1, 1, -1). We use the same definition of symmetrically
opposite vertices in the remaining examples in this paper. For instance,
in example \ref{four-d-cube}, the symmetrically opposite vertex of
(-1, 1, -1, 1) is (1, -1, 1, -1). Therefore the points belonging to
the clusters around these vertices belong to the same class.

We have drawn samples from these clusters to formulate the train data
set, these are 100 sample points randomly generated within each spherical
cluster (100 samples per cluster) and test data set (50 samples per
cluster). A feed-forward neural network was then trained to classify
the training data set. The architecture of the network is as follows:
dimension of the input layer is 3, dimension of the first hidden layer
is 8 (equals the assumed number of planes required to split the clusters, though for this simple case the 3 coordinate planes are sufficient to split the clusters we do not use this information),dimension of the second hidden layer is 8 (equals the number of clusters) and the dimension of the output layer is 4 (equals the number of classes).
Therefore by using the network architecture: 3-8-8-4 The feed forward
neural network was trained using Back propagation algorithm and it
gave 100\% classification accuracy on both the training and test data
sets.

\subsection{Four Dimensional Cube}

\label{four-d-cube} This is a 4-d problem involving a hypercube
which is centered at the origin and whose neighboring vertices at a distance
of 2 apart from each other. For convenience we assume the shape of
the clusters in this and the next problem to be that of a 4d sphere. There
are thus 16 spherical clusters, centered at each of the vertices,
and having a radius of 0.3. The symmetrically opposite clusters of
the 4-d cube belong to the same class, (i.e. if the cluster centered
at the coordinate (1,1,1,1) belongs to class 1, then the cluster whose
center is situated at (-1,-1,-1,-1) also belongs to class 1. Hence,
there are a total of 8 classes.

As in the above experiment, we have drawn samples from these clusters
to formulate the train data set (100 samples per cluster) and test
data set (50 samples per cluster). A feed-forward neural network was
then trained to classify the training data set.

(i) Using the 3 layers of processing elements in the architecture:
The architecture of the network is as follows: dimension of the input
layer is 4, dimension of the first hidden layer is 16 (equals the assumed 
number of planes required to split the clusters), dimension of the
second hidden layer is 16 (equals the number of clusters) and the
dimension of the output layer is 8 (equals the number of classes).

(a) Using: 4-16-16-8 architecture, the Back propagation algorithm
produced 100\% classification accuracy on both the training and test
data sets.

(b) Actually for this problem we can show that just 4 planes are sufficient
to separate the cluster, these are the 4 coordinate planes: x= 0;
y=0; z=0; t=0;

ii) Using the 4 layers of processing elements in the architecture:

(a) 4-16-2-8

(b) 4-15-5-8 (c) 4-9-9-8 The feed forward neural network was trained
using Back propagation algorithm and it gave 100\% classification
accuracy on both the training and test data sets.

\subsection{Nested Four Dimensional Cubes}

Here we consider a 4-d problem of a big hypercube which has smaller
hypercubes centered at each of its 16 vertices. That is each smaller
hypercube has its center at one of the vertices of the larger hypercube.
Thus a total of 256 spherical clusters belonging to 8 classes. The
neighboring vertices of the bigger hypercube are at a distance of
4 apart from each other. The vertices of this bigger 4-d cube form
the center of the smaller 4-d cube. The neighboring vertices of the
smaller hypercube are at a distance of 2 apart from each other. Hence
there are 256 (16x16) clusters having a radius of 0.7 (there are no
clusters at the vertices of the bigger hypercube) Now we classify
each cluster as follows. As in the example \ref{four-d-cube}, each
small hypercube will have 16 clusters and clusters symmetrically opposite
will belong to the same class, thus there are 8 classes for each small
hypercube. As there are 16 small hypercubes there will be 256 clusters
belonging to 8 classes. Note we have imposed a symmetry to our problem
by placing all the cubes in such a manner that the edges of each of
the cubes are parallel to one of the coordinate axis. \emph{This symmetry
has been imposed on this and all the subsequent examples considered in this paper.}

(i) Three layer of Processing elements:  As in the above experiment,
we have drawn samples from these clusters to formulate the train data
set (100 samples per cluster) and test data set (50 samples per cluster).
A feed-forward neural network was then trained to classify the training
data set. The architecture of the network is as follows: dimension
of the input layer is 4, dimension of the first hidden layer is 256
(equals the number of planes, which was purposely chosen very high and equal to the number of clusters), dimension
of the second hidden layer is 256 (equals the number of clusters)
and the dimension of the output layer is 8 (equals the number of classes).
Thus the architectures tried out is: (a) 4-256-256-8, The feed
forward neural network was trained using BP algorithm,
which converged in less than 500 epochs, and it gave 100\% classification
accuracy on both the training and test data sets.We have chosen the
number of planes as 256 (equal to the number of clusters), which is
sufficient to distinguish all the clusters from one another by a 256
dimension orientation vector. Actually, it would not have mattered
even if we had chosen more planes than 256.

However, it may be noticed that because of the symmetry of the configuration, only 12 planes are actually required
to separate all 256 clusters, these planes are: x= 0; y=0; z=0; t=0;
x= 1; y=1; z=1; t=1; x= -1; y= -1; z= -1; t= -1; so an architecture
of the type 4-12-256-8 is theoretically sufficient for this problem,
therefore by using various architectures we got results as per this
table:

(a) 4-12-256-8 : 99.852\% train \& 99.672\% test 

(b) 4-13-256-8: 99.191\% train \& 99.117\% test 

(c) 4-14-256-8: 99.891\% train \& 99.641\% test 

(d) 4-18-256-8 :100\% train \& 99.969\% test 

(ii) Architectures with Four layers of Processing elements which also
worked are:

(e) 4-12-40-300-8 (f) 4-12-100-150-8 (g) 4-12-120-80-8 (h) 4-12-50-50-8

\subsection{Generalization of Nested Cube Problem to n dimensions}

The problem 3 which has clusters of smaller cubes placed at corners
of larger cubes can be generalized to $n-$ dimensions. There will
be a large n-dimension cube with smaller n-dimensional cubes at the
corners: so we have $2^{2n}$clusters and if each pair of ``diagonally
opposite'' clusters in the smaller cube belong to the
same class then there will be $2^{n-1}$ classes. We require only
$3n$ planes to separate the clusters so the minimal neural network
architecture will be : $n-3n-2^{2n}-2^{n-1}$.

This problem is interesting because a Radial Basis Function method
would involve $2^{2n}$ distance measurements to classify a single
sample, where as by this method there are only $3n$ linear equations
to be evaluated to obtain the unique Hamming vector which identifies
the same sample.

\subsection{Generalization to sequence of nested cubes one inside the other}

In fact we can still further generalize the n dimensional example
given in the previous section. Suppose we define the previous example
as level-2 nesting: that is we take a large n-dimensional cube and
place smaller n-dimensional cubes at the vertices, each of these small
cubes have a cluster at its vertex. Now we can consider such level-2
structures placed at the vertices of a still larger n-dimensional
cube we get a level-3 structure (ie level-2 nested cubes at the vertices
of another large cube). So we can go on to get a level-r nested structure.
This level-r nested structure will have $2^{rn}$ clusters belonging
to $2^{n-1}$ classes. It can be shown that such a cluster can be
distinguished by using $(2^{r}-1)n$ planes and thus we will have
a NN architecture: $n-(2^{r}-1)n-2^{rn}-2^{n-1}$.

There are only $(2^{r}-1)n$ linear equations to be evaluated to obtain
the unique Hamming vector which identifies a sample which may belong
to any one of the $2^{rn}$ clusters and finds which of the $2^{n-1}$
classes it belongs to. These $(2^{r}-1)n$ linear evaluations may
be once again compared with $2^{rn}$ distance measurements which
would be necessary, to classify a single sample, if one uses the Radial
Basis Function method.

\section{ Application to Classification Problems}

Now it is probably appropriate to answer the query: What type of patterns
and what type of cluster configurations can be easily classified by
 our method? It would be quite apparent by now that if the patterns
are in clear clusters like those given in Example 1 and 2 (Level 1)then
the problem is completely classifiable by using the above mentioned
neural architecture, the EXAMPLES section clearly illustrate this
(in particular problem 1 and 2, viz the 3-d and 4-d cubes). However
in some other cases wherein we have clusters within clusters, Level
2, (eg. problem 3; the nested 4-d cube); that is when each large cluster contains
sub-clusters, much like a cluster of galaxies each of which is a cluster
of stars, some more investigation needs to be done. For these and
other cases (cluster configurations of Level-r), it is possible to
estimate the number of planes required. However, the precise number
of planes would depend on the number of clusters and their geometrical
positions in feature space. So we can only make an estimate. This
estimate helps determine a possible architecture for the Neural network
classifier.Before proceeding to our estimates, we would first need
a definition.

 \textbf{Malleable cluster:} We will define (consider)
a cluster as 'malleable' if (i) a sample point is classifiable to
a cluster by just taking its Euclidean distance to the centroid of
a cluster, OR (ii) a sample point can be associated to its cluster by
a k nearest neighbor algorithm. All clusters will be assumed to be
malleable. We further assume that different clusters are separable
from one another by planes, if necessary a cluster may be divided
into two or more parts to facilitate such a separation (see figure
\ref{fig:fig-d}).

We wish to say without being \emph{ad nausea} that in cases where clusters belonging to different classes overlap with each other then it is not possible to classify the problem without using probabilistic techniques and we do not consider such situations in the paper. It could also mean that we have not taken
enough number of relevant features to solve the problem.

\textbf{ESTIMATION OF NUMBER OF PLANES}

In the next three subsections, we give methods with heuristic proofs on how to estimate the number planes 
which can separate  Level 1 and Level 2 clusters and also for the case when the clusters are not too sparse.

Though our proofs are heuristic, it may be mentioned that our estimates are in concordance with the bounds proved by Ralph P. Boland and Jorge Urrutia {[}28{]},1995, who in their work had elegantly exploited the crucial fact: In $n$-dimension space a single plane, in general, can simultaneously separate $n$ pairs of points(randomly placed, but not all in the same plane), thus if we choose the first pair of $2n$ points (among N), the first plane thus cuts these $2n$ points and places them into two sets one on either side of the plane,\footnote{Another way of looking at this is to think that each pair of points as a line segment which has a midpoint, since there are $n$ pairs, one can always find the $n$ coefficients $ \alpha_i, (i=1,2,..,n)$ of a plane (say) $ 1 + \alpha_1 x_{1}+ \alpha_2 x_{2}+..+\alpha_n x_{n} = 0 $ which passes through these $n$ midpoints.}  this plane of course divides the other points among  N to to either side; after this $n$ new pairs of $2n$ points are chosen such that each pair is unseparated, a second plane is then chosen which divides the new $n$ pairs and also the space to 4 'quadrants' , the next plane gives 8 `quadrants`,  the process continues and new planes are added, but must quickly end because all the N points will be soon exhausted.\footnote{ Another crucial point to note regarding n-dimensional geometry: Every time you add a plane in $n$-dimensional space you are dividing the space and doubling the number of existing number of `quadrants', but this doubling happens only for the first $n$ planes the $(n+1)^{th}$ plane will not double the `quadrants' but create a `region' confined by other  n planes. Remember we have chosen n large and $N < 2^n$.} The proofs by Boland and Urrutia{[}28{]}are involved though rigorous.

\subsection{Estimate of number of planes: Clusters of Level 1}

 We show that for  problems, involving large n-dimensional feature space, which has N clusters,  $N<2^{n}$,
sparsely and randomly distributed and configured as Level 1, the number of planes
$q$ are $O(log_{2}(N))$.

As is known a 2d space has 4 quadrants,3d space has 4 quadrants and
n dimensional space has $2^{n}$ quadrants. Suppose the dimension
of the feature space is large (say 40), then it is most likely, in
practical situations such as face recognition, disease classification
etc., that the number of clusters (say 10000), will be far less than
the number of quadrants, (as $2^{40}\approx10^{12}$) that is, the
number of clusters will be sparsely and randomly distributed. Therefore
an interesting question arises: Is it possible to transform the feature
space $X$ of n-dimension to another n-dimensional $Z$ space such
that each cluster in $X$ space finds itself to be in one quadrant
in $Z$ space, such that each cluster is in a different coordinate
in this $Z$ space? If this is so then the problem can be tackled
in $Z$ space instead of the original feature space, thus making the
classification problem trivial. The answer to the question is yes,
if the cluster configuration is of type Level-1 In fact, the present
problem is closely related to the problem first dealt with by Johnson
and Lindenstrauss {[}21{]}(1984), who showed that if one is given
N points in a large n dimensional space then it is possible to map
these N points to a lower dimension space k of order $k=log_{2}(N)$,
in such a manner, that the pairwise distances between these points
are approximately preserved, (in fact our requirement is much less
stringent we only require that the `centroid' of N clusters, be mapped
to a different `quadrant'). The transformation is easy: Every point
P, in n dimension space, whose coordinate is $\underline{x}_{P}$
and which belongs to cluster $i$ can be transformed to $\underline{x'}_{P}$
another point in $n$ dimension space. This transformation from $X$
space to $X'$ space is given by : $\underline{x'}_{P}$= $\underline{C}_{i}$
+($\underline{x}_{P}$ - $\underline{x}_{i}$) Where $\underline{x}_{i}$
is the centroid of cluster $i$ in the $X$ space; $\underline{C}_{i}$
is the coordinate of the point to which the centroid of $i$ has been
shifted in $X'$ space. We can choose $\underline{C}_{i}$ to be sufficiently
far away from the origin such that its distance, from the origin of
$X'$ space is larger than the radius of the largest cluster (for
convenience, we can choose the origin of the $X'$ space to be the
global centroid of the sample space). Typically, if $n=5$ we could
choose some point say, $\underline{C}_{i}=D(1,1,-1,1,-1)$ where $D$
is sufficiently large. Thus we see that the problem is classifiable
in $X'$ space, and a classifier with (say) $q$ planes, $q=log_{2}(N)$,
exists and since the transformation from $X$ space to $X'$ space
is essentially linear and the clusters are sparse in $X'$ space and
can be separated by these $q$ planes, then a similar classifier exists
in $x$ space. In $X'$ space the centroid of each cluster can be
given `coordinates' by measuring the perpendicular distance from each
of the $q$ planes to get the `coordinates' $(z1,z2,...zq)$, ie
related to the orientation vector, therefore each cluster will be
in a `quadrant' of the q dimensional $z$ space. A situation somewhat
similar to Johnson and Lindenstrauss {[}21{]}-{[}22{]} because $q=O(log_{2}(N))$.
QED.%
\footnote{ An alternative argument can be had by transforming all the points
in X-space of n dimension to points on the surface of a sphere (radius
R) of n+1 dimensional X'-space. After, choosing the origin as the
global centroid of all the clusters, we use the transformation $x'_{i}=Rx_{i}/A\,(i=1,2,..,n);$
and $x'_{n+1}=R/A$ with the choice $A=(1+x_{1}^{2}+x_{2}^{2}+..x_{n}^{2})^{1/2}$.
The clusters on the sphere can be separated from one another (because
they are sparse) by (say) q `great circles', each contained in a plane
through the origin. We thus arrive at the same result.%
}

 From the above we see that problems involving labeled data in class
1 are always classifiable by transformation into $Z$ space. Thus
we see when the number of clusters N are such that $N<2^{n}$, n being
the dimension of space, then we require only $q$ planes $q=log_{2}(N)$.
In the example we see that Problem 1 and Problem 2 are problems of
class 1 type. further, problem 1 is a cluster in three dimension space
involving 8 clusters in this case 8 is equal to $2^{3}$, hence we
see that the clusters are sparse and hence it can be solved by using
only 3 planes the equations of these 3 planes are x=0, y=0, and z=0.
Similarly in Problem 2 we have a 4 dimension cube involving 16 clusters
which is equal to $2^{4}$, here again we need only 4 planes whose
equations are x=0,y=0,z=0 and t=0. In problem 3 we have too many clusters
(256) which is much more than $2^{4}$. Therefore, we will deal with
this case later.

\subsection{Estimate of number of planes: Clusters of Level 2}

Here again we assume that the number of clusters N, $ N < 2^{n}$. In
this case, there is a large cluster involving 'regions` each of which
consists of clusters (analogous to galaxies and stars). We can solve
this as follows: we divide the problem into K regions such that each
region does not have more than $2^{n}$ clusters. We can separate
these K regions by $ log_2(K) $ planes and each of these have a max of N' $\approx N/K$
clusters in a region by $log_{2}(N')$ planes, thus the total number
of planes $q$, would be $O(K log_{2}(N/K))+ O(log_2(K))$.

We can estimate the number of planes by repeatedly using the logic
of the previous paragraph for clusters of Level-r. In problem 3 we
have too many clusters (256) which is much more than $2^{4}$. These
are 16 large clusters each containing 16 smaller clusters therefore
one would have thought that they would require 16 X 4 + 4 =68 planes,
however from the symmetry of the problem we see that we require only
12 planes. Therefore we see that 3, 4, and 12 planes are sufficient
to solve problems 1, 2 and 3 respectively; these are the minimal.
Now it is interesting to see if the Back-Propagation algorithm can
discover the weights of the planes if the number of planes are
specified, we report that if the cluster sizes are small and if the
separation between clusters is large than the algorithm succeeds,
else more number of planes are required. Details are provided in the
Example section. What happens if we have labeled clusters and if we
do not know the number of clusters? We have seen (in the Example section) that we need to approximately
guess the number of clusters as $m$ and if the problem is of Level-1
than choose the number of planes to be a little higher than $log_{2}(m)$.
In fact if we choose the number of planes is equal to the number of
clusters, the upper limit, the problem is automatically resolved but
this is inefficient. In the Example problems, we did start with this
rather inefficient guess (by assuming the number of planes as equal
to the number of clusters) and apply the BP algorithm. We then solved
the examples by using a variety of more efficient architectures 
which were then trained by the Back Propagation algorithm.

\subsection{Problems when the number of clusters are not sparse}

We had assumed that the number of clusters are sparse. What happens
if the number of clusters are not smaller than $2^{n}$? In this case
we use the method previously try to divide the total number of clusters
as belonging to different regions. Choose K regions, such that each
of the K regions does not have more than $2^{n}$ clusters. Of course
the final answer will depend on the geometrical distribution of the
clusters. See ref {[}28{]} for further details on this subject.

\subsection{Some points for implementation in practical classification problems}

For the sake of completeness, we briefly suggest a means of implementation
of the method of classification described in this paper for practical
cases.

(i) Choosing an Architecture The procedure for software implementation
could be as follows: When data is first given, a suitable nearest
neighbor clustering algorithm may be applied (may be done after suitable
dimension reduction). This will give the number of clusters as shown
in figure \ref{fig:fig-e}. The number of separating planes will be
ascertained or estimated. Normally for sparsely distributed, N, clusters
in high n-dimensional space, the number of planes will be $O(log_{2}(N)$,
in practical cases the number of planes can be taken to be 30\% or
40\% more than $log_{2}(N)$, the exact number of planes are not necessary
because an over specification does not matter, the number of layers
of processing elements will be three, thus the architecture of the
ANN is known as the number of classes is known for a supervised problem.
Then the well-known back propagation (BP) algorithm could be employed using
this chosen architecture to solve the classification problem just
as what was demonstrated in section IV.

(ii) Evaluating a chosen Architecture

Suppose we have two Architectures, which give equally good predictions,how
do we say which is better? One way is to use the concept of Occum's
Razor, in order to do this we could use the following two ratios:
(i) the ratio of the `number of equations'%
\footnote{We define the `number of equations' as the total number of conditions
imposed while training.This is equal to number of training samples
multiplied by the number of processing elements in the last layer.%
} fitted (while training) to the total number of weights used in the
neural architecture, we may call this the knowledge content ratio
per weight (KCR),(ii) the second ratio is nothing but the first multiplied
by the fraction of correct predictions (fcp)on unseen test samples,
this would give the prediction efficiency per unit weight (PEW), $PEW=KCR.fcp$
It is best to use that architecture which has the highest possible
KCR or PEW. 

\begin{table}[htp]
\begin{center}
\begin{tabular}{|c|c|c|c|c|}
\hline 
\textbf{Architecture}  & \textbf{Train Accuracy}  & \textbf{Test Accuracy}  & \textbf{KCR}  & \textbf{PEW} \tabularnewline
\hline 
4-12-256-8  & 99.852\%  & 99.672\%  & 18.81  & 18.74 \tabularnewline
\hline 
4-13-256-8  & 99.191\%  & 99.117\%  & 17.95  & 17.79 \tabularnewline
\hline 
4-14-256-8  & 99.891\%  & 99.641\%  & 17.16  & 17.10 \tabularnewline
\hline 
4-18-256-8  & 100\%  & 99.969\%  & 14.52  & 14.51 \tabularnewline
\hline 
4-256-256-8  & 100\%  & 100\%  & 1.48  & 1.48 \tabularnewline
\hline 
\end{tabular}
\label{table:problem-3-res} 
\end{center}
\end{table}

The Table shown above gives the values of KCR and PEW
for problem 3 (the nested 4 d clusters), for a variety of architectures 
which were trained using BP.

\section{The method of orientation vectors in not NP hard}

Suppose we have arrived at our so called {}``Minimalistic'' architecture
by using the method of Orientation vectors for solving a particular
problem involving N clusters and k classes. (It is assumed that in this section we are dealing with large dimension space with sparse cluster). Now what happens if we
increase the number of clusters by $\Delta(N)$ and the number of
classes from k to k+1? By this time, we have covered enough ground
to be able to answer this question.

Suppose we a have a certain number of clusters say $N=N_f$, in a large $n$ dimension space,
 how do we begin to separate them by planes? We will now describe such a process.
 We start with a certain number of clusters in an initial set (say) $N=N_0$, belonging to the actual configuration of $N=N_f$ clusters and then choose an initial set of planes $q=q_0$, to separate these $N_0$, as a start \footnote{Perhaps a good way to visualize a particular situation, is to assume that $n=50; N_f=10000$ and starting values: $N_0= 10$ and $q_0 =10$, this satisfies $N_0 <<2^{q_0}$.} we will assume $N_0 <<2^{q_0}$.
We will then include more clusters into this set and at the  same time choosing an additional plane (or planes) to separate the new arrivals of clusters from one another and from those which are already in this set. This study will help us to understand that as we increase the number of clusters, the number of planes needs to be increased at a far, far lesser rate, thus demonstrating that our method is not NP hard. We will then arrive at $N=N_f$ and $q=q_f= O(log_2(N_f))$ .
Suppose  that at some intermediate stage of our process, we have arrived at a situation where the clusters are as shown in figure \ref{fig:fig-e} and this stage (say) we have $N=N$ in our set which are all separated by $q$ planes.    Now suppose we include a new cluster from the configurations thus increasing the number of clusters in our collection by one,N to N+1, this new cluster has to appear some where in the diagram. If it is far away from the entire figure say
`above' all the rest then it can be separated from the others by introducing
just one 'horizontal` plane. Now if the new cluster appears inside
the figure, then it will have, at most, one neighboring cluster from
which it is not separated by a plane. This is because if it has (say)
\emph{two} neighbors not separated by a plane, then this implies that
there is no plane separating the other two clusters - an impossibility:
since all the other clusters have already been separated from each
other.(We ignore, the rare case when the new cluster will happen to
cut by a plane, since the clusters are sparsely distributed). Let
us say that this new cluster has a neighbor $4_{3}$, then in this
case adding a new plane, $q \rightarrow q+1$, separates this new cluster from $4_{3}$ and
automatically isolates the new cluster from all the others.

\begin{figure}[htp]
 \begin{center}
\includegraphics[scale=0.35]{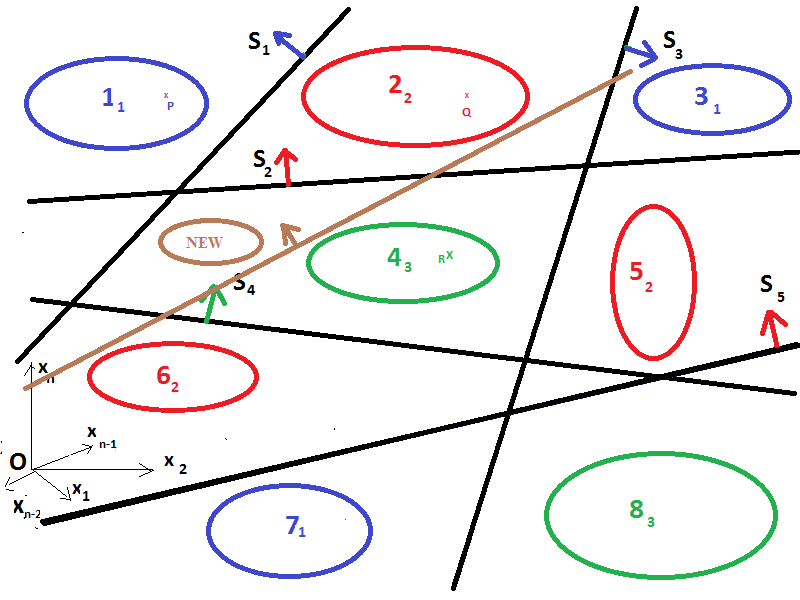} \caption{Cluster of sample points in n-dimensional space}

\label{fig:fig-j}
\end{center}
\end{figure}

$ $

It is worthwhile to investigate a little further: What happens if we now add 
one more cluster to our set the $(N+2)^{nd}$ cluster? The answer to this is not too easy especially if we are dealing with crowded clusters in low dimension spaces. Surprisingly as is shown in Ref [30], it is easier to separate points by planes in a large n-dimension space, rather than clusters. This is because cluster shapes vary - they may be filamentary, dragon like or amoeba like objects and to define the shape of a cluster requires more parameters than to define the coefficients of a plane! In large dimension space the many degrees of freedom available allows one to separate innumerable points with comparatively fewer planes. However for sparse clusters this question  will be answered by following the methods of
Ref[30], but with some what modified arguments because here, we are dealing with separation of sparse clusters by planes rather than separation of points by planes as was done in [30].It will be clear that most of the time
 we need not add one more plane, the new plane can be adjusted so that it can separate both 
the cluster number N+1  and the cluster number N+2 at the same time. Now if the
 $(N+2)^{nd}$ cluster
falls in an empty quadrant among all the $2^{q+1}$ current quadrants then it is already separated from the others
and we don't need to do anything (no need to add a plane); but if it falls in the a quadrant where say 
one of the existing clusters (say) $7_1$ resides then we do as follows: we find the centroid of the $7_1$ and join it to the centroid of this N+2 cluster, these two centroids can be thought of as the ends of a line segment, similarly the 
centroids of the N+1 and $4_3$ can be thought of as the ends of another line segment - we now modify the $q+1$ plane so that it passes through the ``mid points'' of these two segments thus separating the $N+1$ and $N+2$ from all the others and themselves. (Remember $N+1$ and $N+2$ are already separated from each other because it is assumed that they are in different quadrants). If the next $n$ clusters do not fall in empty quadrants, then we can actually add a total of $N+n$ clusters into our set and in all probability the $(q+1)^{st}$ will take care of all of them. Only when we have cluster number $N+n+1$ do we may need to add the $(q+2)^{nd}$ plane. Actually this addition of a new plane maynot even be needed as soon as cluster number $(N+n+1)$ because many of the $n$ new clusters which are added will likely fall on empty quadrants so there is no need to immediately account for them by adjusting the $(q+1)^{st}$ plane or adding the$(q+2)^{nd}$ plane. (In this para we have extended the logic enunciated in para preceding section V A). The logic will fail if the any of the $n$ new clusters happen to fall in the same quadrant or two of them sharing the same quadrant space with an existing cluster say $4_3$, then of course we must introduce the $(q+2)^{nd} $ as soon as this happens. But the chance of two new (given) clusters falling in the same quadrant has a probability of $1/2^{q+1}$ which is very rare indeed! (And if this rare event happens, there is no real problem: we simply add a new plane at this point and start counting the clusters from here on).\footnote{Having arrived at N clusters and q planes we have added the $(q+1)^{st}$ plane, we now have $2^{q+1}$ quadrants from the present $2^q$; it is interesting to conjecture how many clusters can we accomodate before we fill up the $2^{q+1}$ quadrants, or before we need the $(q+2)^{nd}$ plane? The above logic seems to indicate that if our starting $q$ was much larger than the required that is if $N < 2^q$, it is reasonable to expect that this  must be of $O(N)$ clusters , see Ref[30] for more details. Another problem is when the new plane cuts across some existing cluster if this happens (an event not likely for sparse clusters)the resulting cluster needs to be treated as two separate clusters - the clusters need to be renumbered.}   
But we have already shown that the number of planes for sparse clusters
is $log_{2}(N)$, we thus have proved the following: The present method
of classification using `Orientation Vectors', is NOT NP hard for the 
type of problems we are dealing with.\footnote{It will probably be appreciated by those who have followed our logic, closely till now, that the situation is far from NP Hard, in fact, the relationship $q= O(log_2(N))$ greatly underlines the efficiency of the present method.}  If
the number of clusters is increased from $N$ to $N+\Delta N$, then
the number of planes increase at worst linearly by $\Delta N$ and
at best only logarithmically $\Delta(log(N))$. Increasing the number
of classes from k to k+1 only increases a processing elements by one. QED

\textbf{A NOTE : Regarding Algorithm}
It is perhaps quite obvious that the above process of including clusters and determining planes can be the basis of an algorithm, the details of the algorithm is available in {[}29 b{]},{[}30{]}  as applied to separation of points by planes, the methods of which can be modified to separation of sparse clusters by planes. The  results of this study will be reported subsequently.

\subsection{Dimension reduction}

Dimension reduction is done by using auto encoders or by MNN's , which
are nothing but neural networks with many layers and a converging
diverging type of architecture. See figure

The task of the auto encoder is to reduce the dimensionality of input
data and they are trained in such a way that the output just reproduces,
{}``mirrors'', the input. Since an auto encoder first reduces the
data from an initial dimension say $n$ to a lower dimension say $m$
and then increases it back to $n$, albeit in stages, we can think
of an auto encoder (or MNN) as a mapping engine which uses a NN architecture
of a special type. There is an alternative way of training an auto
encoder ie. is by considering it as a Boltzmann machine (with binary
units), however we will not consider this here primarily because a
Boltzmann machine is stochastic and non-deterministic and secondly
its binary nature makes applications more difficult. So we will consider
an auto encoder (or MNN) as a mapping engine which uses a NN having
a special converging-diverging type of architecture.

The purpose of the auto encoder is to reduce the dimension of the
data: If all the samples belong to an $n$-dimensional X-space, it
is assumed that each sample uses more number of dimensions to describe
an exemplar then strictly necessary: hence all the $n$-components
of an input vector (sample) are not really all independent and it
is hypothesized that the data really can be described by say$m$ variables
Therefore for every input sample $(x_{1,}x_{2,}..,x_{n})$ in X-space
there exists some $m$ dimensional vector say $(y_{1,}y_{2,}..,y_{m})$
in Y-space, which describes the input vector. If such an equivalence
between every vector in X-space and some vector in Y-space exists
for all exemplars in the entire data set in X-space, then we can conclude
that all the data can be described by using only m-dimensions. We
can then say that every sample in X-space is a function of $m-$variables
in Y-space and we can think of the components of $m$-variables in
Y-space as independent variables and the $n$-components of the X-space
vectors are actually not independent but dependent on these $m$ variables.
The function of an auto encoder (MNN) is to capture this functional
dependence between X and Y, and a neural architecture is used for
approximately capturing this functional dependence. The weights of
the processing elements are determined by imposing the condition that
it should ``mirror'' each input sample in the data set. The figure
below shows a sequence of transformations X-space to S-space to T-space
then to Y-space then to R-space to U-space and back to X-space (since
the condition V=X is imposed). These can be thought of as a sequence
of mappings (functions of functions etc.) which reduces the data from 
$n$-dimension X-space to $m$-dimension Y-space and back to $n$-dimension
X-space (V-space), in stages: $n>k>i>m<j<l<n$ . The variables $(y_{1,}y_{2,}..,y_{m})$
can be considered as the reduced ``independent'' variables
\footnote{Think of a sample vector (point) in X space as a large sized photograph
involving $n$ pixels and its reduced sized photograph of $m$ pixels
as a vector in Y space representing the same photograph, $m<n,$ in
addition we may have to think of $m$ as a measure of the smallest
sized photograph (the smallest number $m$), which can be used to
distinguish the photographs in the input set one from another.}
and the input data as n-dimensional dependent variables in x space.

\begin{figure}[htp]
 \begin{center}
\includegraphics[scale=0.25]{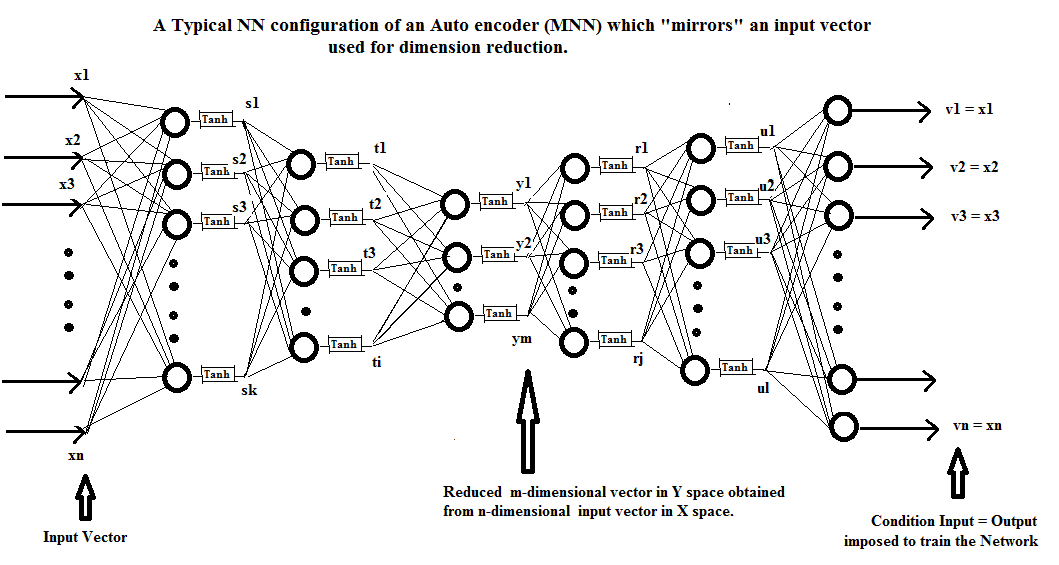} \caption{A Typical Configuration of an Auto encoder (MNN)}

\label{fig:fig-k} 
\end{center}
\end{figure}

The purpose of this section is just to demonstrate only two facts:

(i) That the mapping performed by a fully trained auto encoder (MNN)
is such that each cluster starting from a cluster in an input layer
is mapped to a unique cluster in the next layer,and this is true from
layer to layer. To prove this we take a simpler MNN shown below:

\begin{figure}[htp]
\begin{center}
\includegraphics[scale=0.25]{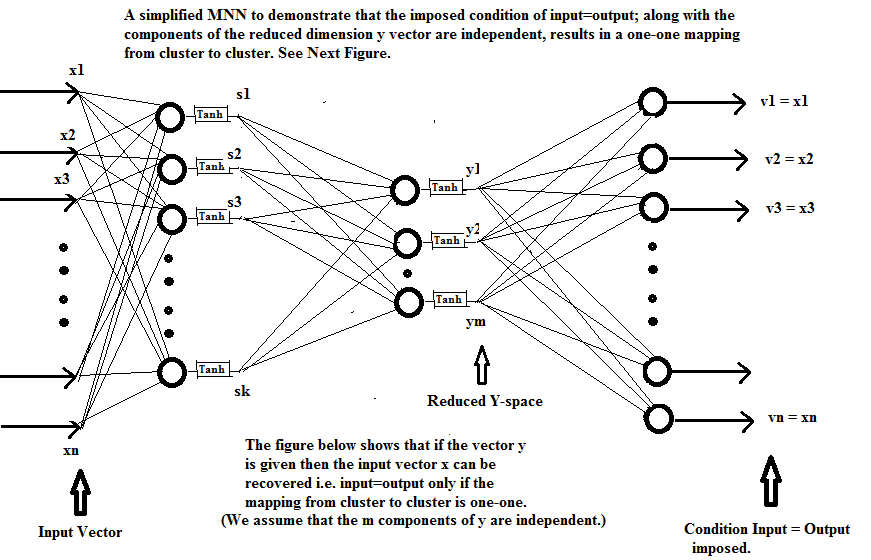} \caption{A Simple Auto encoder (MNN)}

\label{fig:fig-l}
\end{center}
\end{figure}

The mappings made by the above architecture are shown pictorially
below: 
\begin{figure}[htp]
 \begin{center}
\includegraphics[scale=0.25]{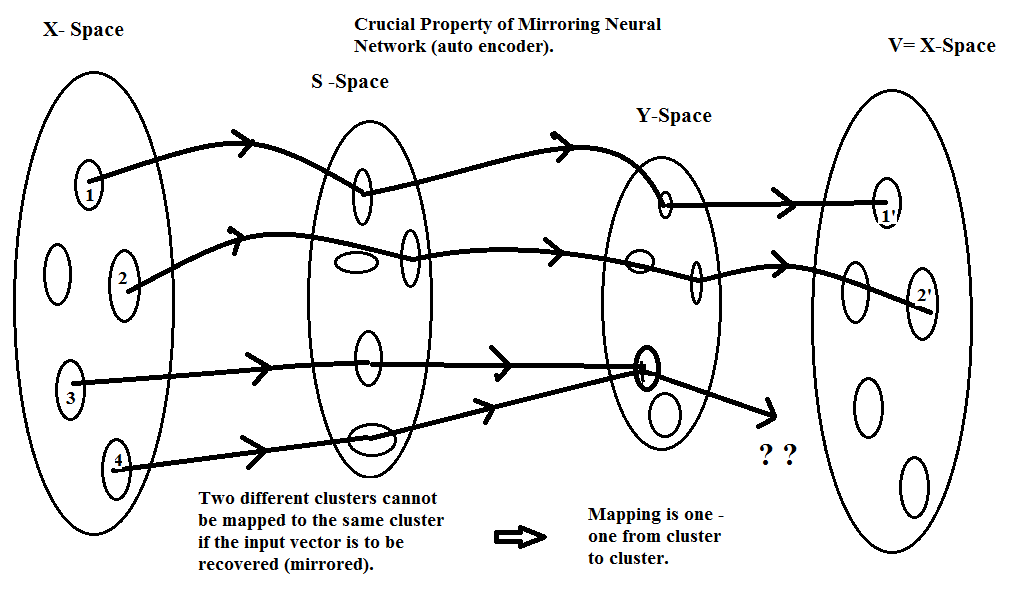} \caption{Mapping Property of Auto encoder (MNN)}

\label{fig:fig-l} 
\end{center}
\end{figure}

It is clear that since we want the MNN to {}``mirror'' each vector
this property of the mapping moving from one cluster to another is
correctly depicted for points 1 and 2 as shown above. The case of
points 3 and 4 which start from different clusters and land in the
same cluster cannot happen, because if the points 3 and 4 land up
in the same cluster (as shown for e.g. in Y-space), the network cannot
{}``mirror'' the input thus the input vector cannot be recovered.
This property is important because it leads to an important Theorem
which shows that such MNN architectures can be used for hierarchical
classifiers where classes can be further sub classified to subclasses.See Ref. {[}24{]}-{[}25{]}. 

(ii) The second fact is with regard to the
activation function introduced after each processing element. Going
back to Fig 2, we have defined $s_{1}=tanh(\beta y_{1})$, now if
$\beta$ is large say $\beta=5$ then the output $s_{1}$ becomes
either close to +1 or -1 , hence if we choose such a $\beta$ for
all the processing elements in the network, then all the clusters
will be mapped near the center of each quadrant in each space. Such
a situation makes the training of a MNN difficult, so if one chooses
the activation functions such that $\beta\approx0.5$ or smaller than
the mapping will take place in such a manner that the image points
fill up the quadrant space and not just crowd around its `center',
this situation makes the training easier and it makes the functioning
of the architecture more flexible and a suitable configuration that
reduces the input data can be found more easily. Many researchers
in Deep Learning have found this to be the case in their numerical
experiments.

\section{Conclusion}

To conclude, we have made the following contributions in this paper: 
\begin{itemize}
\item We have introduced the method of Orientation Vectors to show that
the classification problem using neural networks can be solved in
a manner which is NOT NP hard. 
\item We have shown a correspondence between our method of Orientation Vectors
and the Kolmogorov technique provided some stringent conditions in
the latter are relaxed. 
\item We have shown proved that a classification problem wherein each cluster
is distinguishable from the other, is always solvable (classifiable)
with a suitable feed forward neural network architecture containing
three hidden layers. 
\item The number of processing elements solely depends on the number of
clusters in the feature space, 
\item Further, we have shown when the feature space is of large$n$ dimension
and the number of clusters, $N$, are sparse s.t. $N<2^{n}$, then
the processing elements in the first layer are $O(log_{2}(N))$. 
\item When the problem size increases that is if the number of clusters
is increased from $N$ to $N+\Delta N$, then the number of planes
increase at worst, linearly by $\Delta N$ , and at best, only logarithmically
by $\Delta(log(N))$. Increasing the number of classes from $k$ to
$k+1$ only increases the processing elements by one. 
\item Many examples have been explicitly solved and it has been demonstrated
through them that the method of Orientation Vectors requires much
less computational effort than Radial Basis Function methods and other
techniques wherein distance computations are required (e.g. statistical). 
\item A practical method of applying the concept of Occum's razor to choose
between two architectures which solves the same classification problem
has been illustrated. 
\item The ramifications of the above findings on the field of Deep Learning
have also been briefly investigated and we have found that it directly
leads to the existence of certain types of NN architectures which
can be used as a {}``mapping engine'', which has the property of
{}``invertibility'', thus improving the prospect of their deployment
for solving problems involving Deep Learning and hierarchical classification.
The latter possibility has a lot of future scope. 
\end{itemize}
As a future work to this paper, we would focus on finding methods
to apply these methods on practical data sets which occur in the areas
of Deep Learning ({[}17{]} - {[}19{]}), {[}31{]} on cloud computing platforms.

\section{ACKNOWLEDGEMENTS}
  The authors thank C Chaitanya of Ozonetel for the many technical discussions that we had with him. We thank the management of SNIST and ALPES for their encouragement.
  
\section{DEDICATION}
   This paper is dedicated to the memory of D.S.M. Vishnu, (1925-2015), Chief Research Engineer, Corporate Research Division BHEL, Vikasnagar Hyderabad.

\section{References}

{[}1{]} A. N. Kolmogorov: On the representation of continuous functions
of many variables by superpositions of continuous functions of one
variable and addition. Doklay Akademii Nauk USSR, 14(5):953 - 956,
(1957). Translated in: Amer. Math Soc. Transl. 28, 55-59 (1963).

{[}2{]} G.G. Lorentz: Approximation of functions. Athena Series, Selected
Topics in Mathematics. Holt, Rinehart, Winston, Inc., New York (1966).

{[}3{]} G.G. Lorentz: The 13th Problem of Hilbert, In Mathematical
Developments arising out of Hilberts Problems, F.E. Browder (ed), Proc. of
Symp. AMS 28, 419-430 (1976).

{[}4{]} G. Lorentz, M. Golitschek, and Y. Makovoz: Constructive Approximation:
Advanced Problems. Springer (1996).

{[}5{]} D. A. Sprecher: On the structure of continuous functions of
several variables. Transactions Amer. Math. Soc, 115(3):340 - 355
(1965).

{[}6{]} D. A. Sprecher: An improvement in the superposition theorem
of Kolmogorov. Journal of Mathematical Analysis and Applications,
38:208 - 213 (1972).

{[}7{]} Bunpei Irie and Sei Miyake: Capabilities of Three-layered
Perceptrons, IEEE International Conference on Neural Networks , pp641-648,
Vol 1.24-27-July, (1988).

{[}8{]} D. A. Sprecher: A numerical implementation of Kolmogorov's
superpositions. Neural Networks, 9(5):765 - 772 (1996).

{[}9{]} D. A. Sprecher: A numerical implementation of Kolmogorov's
superpositions II. Neural Networks, 10(3):447 - 457 (1997).

{[}10{]} Paul C. Kainen and V·era Kurkova: An Integral Upper Bound
for Neural Network Approximation, Neural Computation, 21, 2970-2989
(2009).

{[}11{]} Jürgen Braun, Michael Griebel: On a Constructive Proof of
Kolmogorov's Superposition Theorem, Constructive Approximation,Volume
30, Issue 3, pp 653-675 (2009).

{[}12{]} David Sprecher: On computational algorithms for real-valued
continuous functions of several variables, Neural Networks 59, 16-22(2014).

{[}13{]} Vasco Brattka : From Hilbert's 13th Problem to the theory
of neural networks: constructive aspects of Kolmogorov's Superposition
Theorem, Kolmogrov's Heritage in Mathematics, pp 273-274, Springer
(2007).

{[}14{]} Hecht-Nielsen, R.: Neurocomputing. Addison-Wesley, Reading
(1990).

{[}15{]} Hecht-Nielsen, R.: Kolmogorov's mapping neural network existence
theorem. In Proceedings IEEE International Conference On Neural Networks,
volume II, pages 11-13, New York,IEEE Press (1987).

{[}16{]} Rumelhart, D. E., Hinton, G. E., and R. J. Williams: Learning
representations by back-propagating errors. Nature, 323, 533--536
(1986).

{[}17{]} Yoshua Bengio: Learning Deep Architectures for AI. Foundations
and Trends in Machine Learning: Vol. 2: No. 1, pp 1-127 (2009).

{[}18{]} J. Schmidhuber: Deep Learning in Neural Networks: An Overview.
75 pages,  http:\// arxiv.org/abs/1404.7828,(2014).

{[}19{]} D. George and J.C. Hawkins: Trainable hierarchical memory
system and method, January 24 2012. URL https:\// www.google.com \/ patents\/ US8103603.
US Patent 8,103,603.

{[}20{]} Corrinna Cortes and Vladmir Vapnik: Support-Vector Networks,
Machine Learning, 20, 273-297 (1995)

{[}21{]} William B. Johnson and Joram Lindenstrauss: Extensions of
Lipschitz mappings on to a Hilbert Space, Contemporary Mathematics,
26, pp 189-206 (1984)

{[}22{]}Sanjoy Dasgupta and Anupam Gupta: An Elementary Proof of a
Theorem of Johnson and Lindenstrauss, Random Struct.Alg., 22: 60\textendash{}65,
2002 Wiley Periodicals.

{[}23{]} G.E. Hinton and R.R. Salkhutdinov: Reducing the Dimensionality
with Neural Networks,v 313, Science, pp 504- 507 (2006)

{[}24{]} Dasika Ratna Deepthi and K. Eswaran: A mirroring theorem
and its application to a new method of unsupervised hierarchical pattern
classification. International Journal of Computer Science and Information
Security, pp. 016-025, vol 6, 2009.

{[}25{]} Dasika Ratna Deepthi and K. Eswaran: Pattern recognition
and memory mapping using mirroring neural networks. International
Journal of Computer Applications 1(12):88-96, February 2010.

{[}26{]} K Eswaran: Numenta lightning talk on dimension reduction
and unsupervised learning. In Numenta HTM Workshop, Jun, pages 23-24,
2008a.

{[}27{]} R.P. Lippmann: An introduction to computing with neural nets, IEEE,ASSP magazine, pp 4-22 (1987) 

{[}28{]} Ralph P. Boland and Jorge Urrutia: Separating Collection of points in Euclidean Spaces, Information Processing Letters, vol 53, no.4, pp, 177-183 (1995)

{[}29{]} K.Eswaran:A system and method of classification etc.  Patents filed IPO No.(a) 1256/CHE July 2006 and (b) 2669/CHE June 2015 

{[}30{]} K.Eswaran: A non iterative method of separation of points by planes and its application, Sent for publ. (2015) 
http://arxiv.org/abs/1509.08742

{[}31{]} K.Eswaran and C. Chaitanya: Cloud based unsupervised learning architecture, Recent researches In AI and 
and Knowledge Engg. Data Bases, WSEAS Conf. at Cambridge Univ. U.K. ISBN 978-960-474-273-8, 2011.

\end{document}